\documentclass{article}
\usepackage{ijcai23}

\usepackage{times}
\usepackage{soul}
\usepackage{url}
\usepackage[hidelinks]{hyperref}
\usepackage[utf8]{inputenc}
\usepackage[small]{caption}
\usepackage{graphicx}
\usepackage{amsmath}
\usepackage{amsthm}
\usepackage{booktabs}
\usepackage{algorithm}
\usepackage{algorithmic}
\usepackage[switch]{lineno}

\usepackage{lineno,hyperref}
\usepackage{booktabs}
\usepackage{multirow}
\usepackage{graphicx}
\usepackage{amsfonts,bm}
\usepackage{amssymb}
\usepackage{enumerate}
\usepackage{tabularx}
\usepackage{setspace}
\usepackage{mathrsfs}
\usepackage{diagbox}
\usepackage{amsmath}
\usepackage{colortbl}
\usepackage{bbding}

\title{Dual-view Correlation Hybrid Attention Network for Robust Holistic Mammogram Classification}

\author{
Zhiwei Wang$^{1,2}$ \and
Junlin Xian$^{3}$ \and
Kangyi Liu$^{3}$ \and
Xin Li$^{1,2}$ \and
Qiang Li$^{1,2}$ \And
Xin Yang$^{3}$
\affiliations
$^1$ Britton Chance Center for
Biomedical Photonics, Wuhan National Laboratory for Optoelectronics, Huazhong University of Science and Technology\\
$^2$ MoE Key Laboratory for Biomedical Photonics, Collaborative Innovation Center for Biomedical Engineering, School of Engineering Sciences, Huazhong University of Science and Technology\\
$^3$ School of Electronic Information and Communications, Huazhong University of Science and Technology\\
\emails
\{zwwang, xianjunlin, m20202099, lixin$\_$bme, liqiang8, xinyang2014\}@hust.edu.cn
}

\begin{document}
\maketitle
\begin{abstract}
Mammogram image is important for breast cancer screening, and typically obtained in a dual-view form, i.e., cranio-caudal (CC) and mediolateral oblique (MLO), to provide complementary information.
However, previous methods mostly learn features from the two views independently, which violates the clinical knowledge and ignores the importance of dual-view correlation.
In this paper, we propose a dual-view correlation hybrid attention network (DCHA-Net) for robust holistic mammogram classification.
Specifically, DCHA-Net is carefully designed to extract and reinvent deep features for the two views, and meanwhile to maximize the underlying correlations between them.
A hybrid attention module, consisting of local relation and non-local attention blocks, is proposed to alleviate the spatial misalignment of the paired views in the correlation maximization.
A dual-view correlation loss is introduced to maximize the feature similarity between corresponding strip-like regions with equal distance to the chest wall, motivated by the fact that their features represent the same breast tissues, and thus should be highly-correlated.
Experimental results on two public datasets, i.e., INbreast and CBIS-DDSM, demonstrate that DCHA-Net can well preserve and maximize feature correlations across views, and thus outperforms the state-of-the-arts for classifying a whole mammogram as malignant or not. 
\end{abstract}

\section{Introduction}
Breast cancer is the most common malignant tumor of middle-aged and elderly women, and around 1.2 million women are diagnosed with breast cancer every year~\cite{hosseini2016early}. 
The classification of breast cancer mainly relies on the immunohistochemical diagnosis of breast cancer tissue, which is complex and traumatic and thus can not meet the needs of accurate diagnosis and personalized treatment. 
With the development of medical imaging techniques, radiomics-based breast cancer diagnosis~\cite{lu2018machine} has become a new non-invasive cancer assessment approach, which is comprehensive, easy to obtain and economic.

Among the imaging modalities, mammography has been proven to be effective in early detection and diagnosis~\cite{moss2012impact}.
In a standard mammographic screening examination, the 3D breast will be projected onto a 2D X-ray film. 
Typically, each breast will be exposure in two different angles, i.e., cranio-caudal (CC) view where X-ray from top to bottom and mediolateral oblique (MLO) view where X-ray projects outward and downward at 45 degrees from the inner and upper part of the breast.
Such dual-view mammogram is necessary and sufficient for radiologists to fully understand the 3D breast with 2D X-ray images, and thus can give accurate clinical decisions by following the standard Breast Imaging Reporting and Data System (BI-RADS)~\cite{liberman2002breast}. 
However, mammogram inspecting is time-consuming and expertise-required, and usually suffers from intra- and inter-observer bias~\cite{bae2014breast}.
Therefore, varied computer-aided diagnosis (CAD) systems have been emerging in recent years to provide fast and objective clinical decisions to assist large-scale screening.

Most traditional CADs~\cite{el2002support,arevalo2016representation} for breast cancer screening rely on the analysis of individual lesions, and share a common pipeline mainly consisting of three consecutive steps (i.e., lesion detection, feature extraction, classification).
They usually require a costly manual labeling of lesion masks either in training or test phase, heavily preventing the screening for a large population.
In comparison, several advanced CADs have been proposed to classify a whole mammogram in a \emph{holistic} fashion with only needs of the image-level supervisions indicating whether a mammogram contains malignant lesions or not.
For exampel, Zhu~\emph{et al.} proposed the Deep MIL~\cite{zhu2017deep} to form the holistic mammogram classification problem into a multiple instance learning (MIL) task, and used three different MIL losses, i.e., max pooling loss, label assign loss, and sparsity loss, to fine-tune a pre-trained AlexNet.
Similarly, Shu~\emph{et al.}~\cite{shu2020deep} designed region-based group-max pooling (RGP) and global-based group-max pooling (GGP) to select more representative local features for the holistic mammogram classification.

Despite their success, these CADs can all be categorized as the single-view based approach, which treats CC and MLO views independently.
However, the dual-view mammogram is naturally more suitable and useful than one-view information for reliable diagnosis.
In a clinical practice, radiologists often resort to the MLO view to confirm the suspect lesions found in the CC view.
Furthermore, a standard imaging routine typically provides a paired CC and MLO views for screening, bringing little extra workload for data acquisition to develop dual-view CADs.
In view of these, many dual-view based CADs 
~\cite{alghamdi2021dv,yan2021deep,xian2021towards,cao2021supervised} have been emerged in the last decade.

Most existing dual-view based CADs make use of dual-view information to boost the performance of lesion detection, which is an intermediate task of breast cancer screening.
For example, Yan~\emph{et al.}~\cite{yan2021towards} utilized a shared YOLOv3~\cite{redmon2018yolov3} for proposing mass candidates and then paired each candidate across views and concatenated their features to directly classify whether they are matched or not by a metric network.
Ma~\emph{et al.}~\cite{ma2021cross} proposed the Cross-View Relation Region-based CNN (CVR-RCNN) for robust mass detection by relating each candidate in one view to all candidates in another based on their feature similarities in order to better suppress false detections.
For the task of classification, Bekker~\emph{et al.}~\cite{bekker2015multi} demonstrated a promising improvement for classifying lesions of clustered microcalcification (MC) by combining the results of two single-view logistic regressions on CC and MLO respectively.
Carneiro~\emph{et al.}~\cite{carneiro2015unregistered,carneiro2017automated} combined images from both views associating with their corresponding lesion masks, i.e., mass and MCs, and explored how to fuse and where to fuse those dual-view information in a dual-path CNN.

Although the above dual-view CADs benefit from auxiliary information brought by additional views, they mostly learn features for different views independently, and produce the clinical decision by a simple combination of them (e.g., adding or concatenating).
With no specific constraints, the underlying feature correlations (i.e., consistency and complementarity) across views are often ignored or failed to be captured, which leaves a great improving room.
Furthermore, the holistic mammogram classification remains unstudied, and these CADs all require a prior of lesion masks, seriously hindering the application in large-scale screening.

\begin{figure*}[tbp]	
	\centering	
	\includegraphics[width=0.95\linewidth]{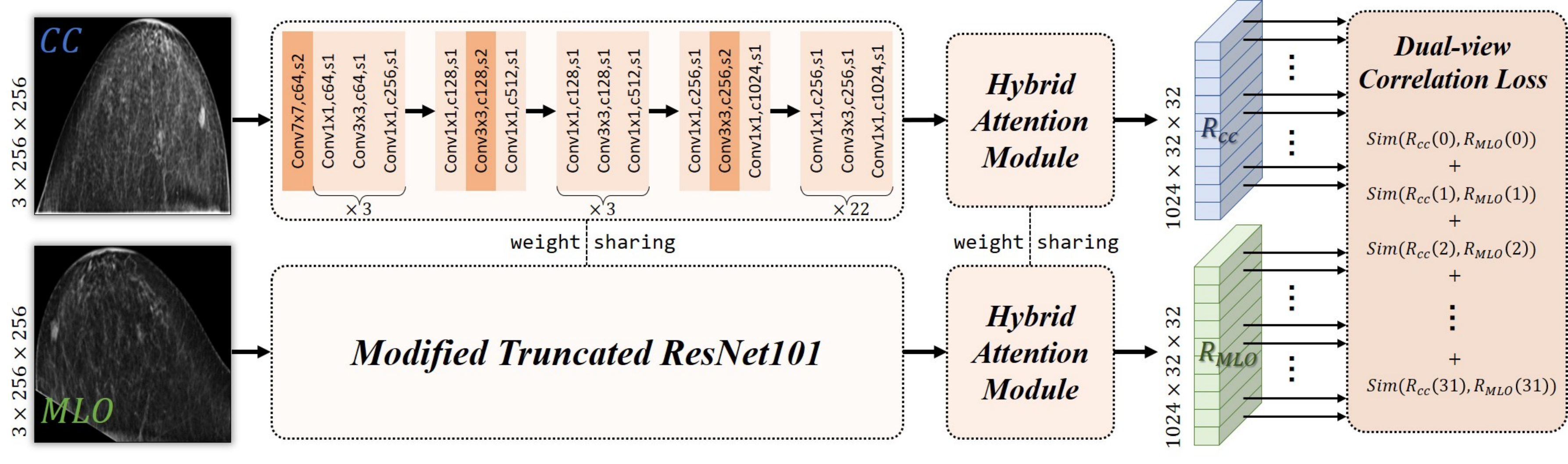}\caption{The framework of our proposed DCHA-Net which utilizes a shared modified truncated ResNet101 (without drawing the skip connections) for feature extraction and a shared hybrid attention module to reinvent feature maps for dual-view correlation maximization.$convK \times K,cN,sM$ means convolving by N kernels with the size of $K \times K$ and stride of $M$.}
	\label{fig:1}	
\end{figure*}

In this paper, we for the first time aim to explicitly maximize the feature correlation across views for robust holistic mammogram classification requiring no lesion masks.
The naive idea is to utilize a shared convolutional neural network (CNN) to extract single-view feature maps in parallel, and maximize the correlation loss proposed in~\cite{yao2017deep} to force the feature maps to be consensus across views.
However, such correlation maximization has a \emph{prerequisite} that the two input images are spatially aligned.
For mammograms in our case, view changing and tissue superimposition make the dual-view images hardly meet the prerequisite, nullifying the benefits of the correlation maximization consequently.

To address this, there are two simple solutions.
One is to spatially align pixels in the CC and MLO views, but practically infeasible.
Another is to reduce spatial dimensions to loosen the requirement of alignment, but inevitably causes non-trivial information loss.
Between these two solutions, we innovatively find a compromise that having each pixel enriched with information from its neighbors, which we argue is equivalent to local spatial dimension reduction but without much information loss.
To this end, we empower the shared CNN by introducing both non-local and local attention mechanisms, and thus name it Dual-view Correlation Hybrid Attention Network (DCHA-Net)\footnote{https://github.com/BryantGary/IJCAI23-Dual-view-Correlation-Hybrid-Attention-Network.}.

Concretely, the DCHA-Net has two shared branches for CC and MLO views respectively and each branch is a modified truncated ResNet101~\cite{he2016deep} with the last few bottlenecks replaced by the proposed hybrid attention module to reinvent features for the purpose of correlation maximization.
The hybrid attention module consists of a local relation block~\cite{hu2019local} and a non-local attention block~\cite{wang2018non} to have each pixel in the resulting feature map contain information from its surroundings (local relation) as well as information of other pixels within its belonging strip-like region parallel to the chest wall (non-local attention).
The motivation is based on a physical fact that two strip-like regions at the same distance from the chest wall are from the same tissue slice and thus matched and high-correlated, that is, \emph{the CC and MLO views are roughly aligned along the direction perpendicular to the chest wall}.
In view of this, the correlation loss is calculated within every matched strip-like regions in CC and MLO views, and optimized to make the two branches of DCHA-Net mutually assist each other.

In summary, our contributions are listed:
\begin{enumerate}
    \item[-] We for the first time propose to learn dual-view features of mammograms by explicitly maximizing the correlations between those matched strip-like regions across views. With such constraint, the consistent and complementary dual-view features could be better captured even under no supervision of lesion labels, yielding a robust performance of holistic mammogram classification.

    \item[-] We propose a DCHA-Net where the hybrid attention module enriches each pixel with its local contexts and global information of its belonging strip-like region, making the correlation maximization correct and effective even if the paired views are not aligned.

    \item[-] We evaluate the proposed DCHA-Net on two well-known datasets, i.e., INbreast~\cite{lee2017curated} and CBIS-DDSM~\cite{moreira2012inbreast}, and the experimental results verify our superior performance over other state-of-the-art methods of breast cancer classification.
\end{enumerate}

\section{Method}
This section is organized as follows: we first describe the framework of DCHA-Net and how to maximize dual-view correlations in Sec.~\ref{sec:ii-a}, then explain two naive solutions to meet the requirement of correlation maximization and lead out our solution in Sec.~\ref{sec:ii-b}.
At last, we detail the hybrid attention module in Sec.~\ref{sec:ii-c}.

\subsection{DCHA-Net and Correlation Maximization}
\label{sec:ii-a}

Fig.~\ref{fig:1} visualizes DCHA-Net, which contains two shared branches for CC and MLO view images, respectively.
Given two images, i.e., $I_{CC}$ and $I_{MLO}$, we first resize them to $256 \times 256$, remove backgrounds and pectoralis muscles, and then align the chest wall with the bottom edge of the image.

In each branch, we utilize a modified and truncated ResNet101~\cite{he2016deep} as a feature extractor, which is detailed in the light orange dashed box of Fig.~\ref{fig:1}.
Compared to the vanilla ResNet101, we abandon the first max pooling layer and the last three bottlenecks (9 layers) to preserve more spatial information and to compromise computational costs for the usage of our proposed hybrid attention module.
By three downsampling layers with the stride of 2, the feature extractor downscales the image by $8^2$ times, yielding feature maps $F_{CC}$ and $F_{MLO}$ with the size of $32 \times 32$.
The feature map is then reinvented by our proposed hybrid attention module to a new feature map, that is, $F_{CC} \rightarrow R_{CC}$ and $F_{MLO} \rightarrow R_{MLO}$, (see Sec.~\ref{sec:ii-c} for details).

\begin{figure}[tbp]	
	\centering	
	\includegraphics[width=0.8\linewidth]{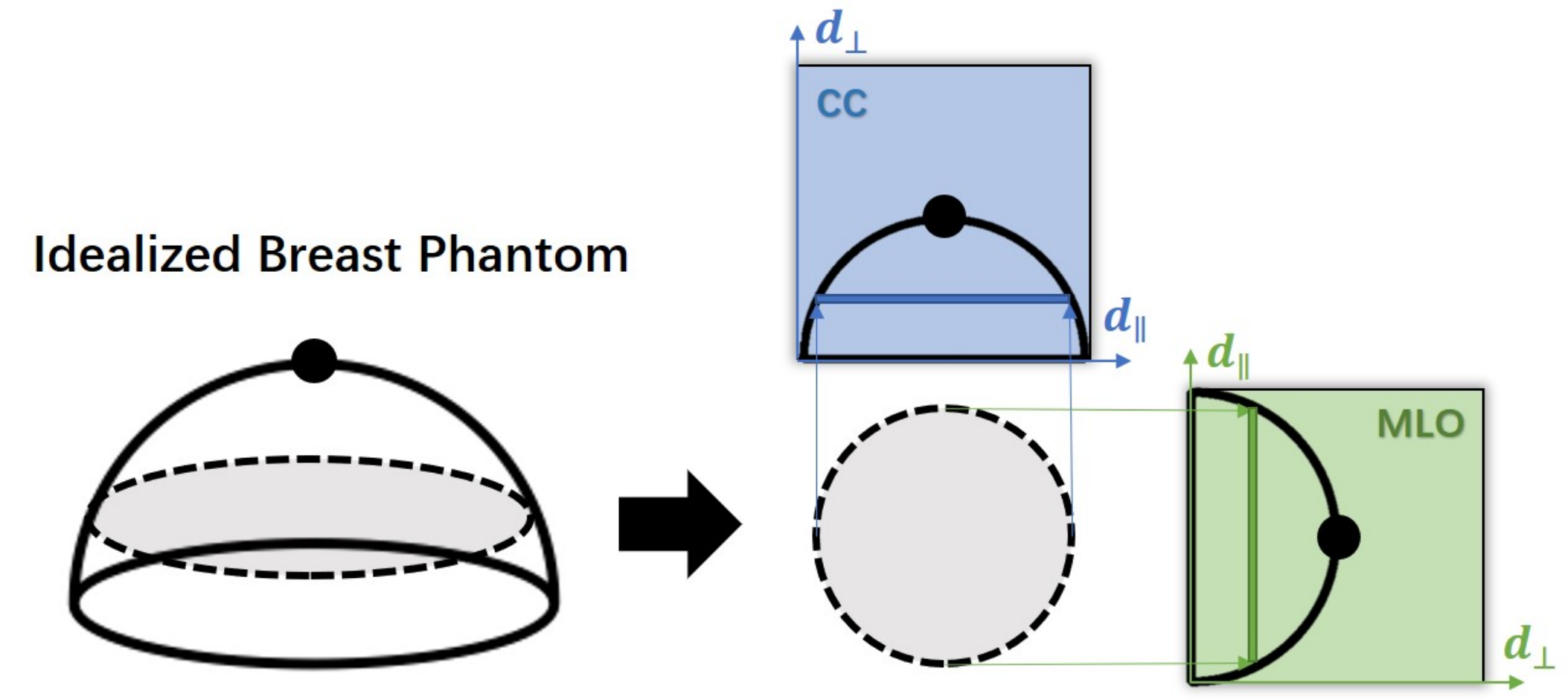}\caption{An idealized model explaining the relationship between 3D breast and 2D dual-view images. $\bm{d}_{\Vert}$ and $\bm{d}_{\perp}$ represent two directions parallel and perpendicular to the chest wall respectively.}
	\label{fig:2}	
\end{figure}

A dual-view correlation loss inspired by~\cite{yao2017deep} is employed to explicitly maximize the feature correlations between the paired feature maps.
Viewing the breast as a rigid semi-sphere, each slice in the 3D breast corresponds to two strip-like regions in both CC and MLO with equal distance to the chest wall as shown in Fig.~\ref{fig:2}, and the matched strip-like regions across views are thus highly-correlated with each other.
Therefore, we propose to maximize correlations between every two row vectors with identical indexes, i.e., $R_{CC}(i)$ and $R_{MLO}(i)$, since their receptive fields just fit two matched strip-like regions.
Concretely, the dual-view correlation loss is calculated as an average of cosine similarities between every matched row vectors in dual-view feature maps: 
\begin{equation}
	\label{eq:1}
	L_{corr}=-\frac{1}{32} \sum^{31}_{i=0} Sim\left( R_{CC}(i),R_{MLO}(i) \right)
\end{equation}
where $R_{CC}(i)$ indicates the $i$-th row vector in $R_{CC}$ and similar to $R_{MLO}(i)$.
The cosine similarity $Sim\left( X,Y \right)$ is calculated as follows:
\begin{equation}
	\label{eq:2}
	Sim\left( X,Y \right)=\frac{\left( X-\overline{X} \right) \left( Y-\overline{Y} \right)}{\lVert  X-\overline{X} \rVert \lVert Y-\overline{Y} \rVert }
\end{equation}
where $\overline{X}$ is a scalar by averaging $X$ and similar to $\overline{Y}$.

Note that the dual-view correlation loss in Eq.~(\ref{eq:1}) is computed based on $R_{CC}$ and $R_{MLO}$ rather than the original feature maps $F_{CC}$ and $F_{MLO}$ because pixels are not aligned across views for the soft and non-rigid breast in the real world, not an idealized phantom shown in Fig.~\ref{fig:2}.
Hence, forcibly calculating $Sim(F_{CC}, F_{MLO})$ could mess up the dual-view correlation loss and make it unable to give full effect.
That is the very reason we introduce the hybrid attention module for feature reinvention.
Before giving details of the hybrid attention module, we highlight our motivations in the next subsection. 

\subsection{Motivation of Hybrid Attention}
\label{sec:ii-b}

The correlation loss is meant to enhance the feature learning for the multi-phase~\cite{zhou2019hyper} or multi-modality~\cite{yao2017deep} data, while remains under-studied for the multi-view mammograms in our case.
The main resistance is that the correlation loss asks the inputs should be spatially aligned beforehand, that is, $X(i,j)$ and $Y(i,j)$ in Eq.~\ref{eq:2} should correspond to the same tissue for the same location $(i, j)$.
To this end, the very straightforward solution is to perform \emph{registration} on dual-view images, which, however, is infeasible since  it is an ill-posed problem to completely disentangle those superimposed tissues from only two X-ray images according to the radon transform theory~\cite{helgason1980radon}.

Another naive solution is \emph{spatial dimension reduction}, performing global average pooling on $F_{CC}$ and $F_{MLO}$, which makes correlation only rely on information along channels.
However, this could result in dramatic information loss, since spatial dimension reduction is equivalent to eliminating feature differences between pixels at different locations.

\begin{figure}[tbp]	
	\centering	
	\includegraphics[width=0.7\linewidth]{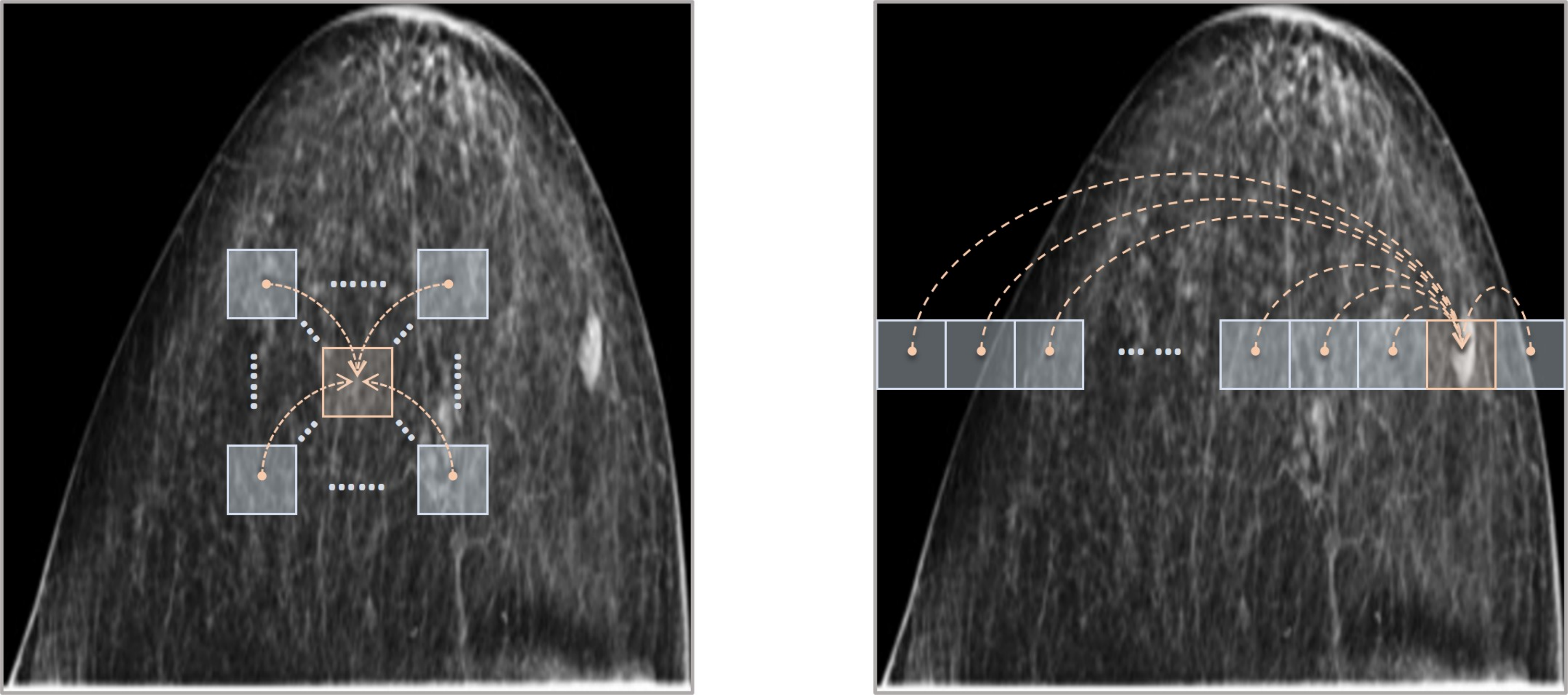}
	\caption{Diagrams of local (left) and non-local (right) attention mechanisms on a CC view image.}
	\label{fig:3}	
\end{figure}

Between these two simple solutions, we come up with a trade-off to alleviate the misalignment problem in dual-view mammograms.
Instead of simply reducing the spatial dimensions, we introduce a local attention mechanism to make each pixel perceive its neighbors within a certain range, as shown in the left image of Fig.~\ref{fig:3}.
Within the range, the misalignment has a chance to be corrected.
That is, if it has a high possibility that a pixel $F_{MLO}(i,j)$ corresponds to its counterpart around $F_{CC}(i,j)$, we relate it to its belonging local patch $\{ F_{CC}(i+\theta,j+\theta), -\sigma < \theta < \sigma, \sigma>0 \}$, where $\sigma$ is the misalignment range, yielding a reinvented map $R_{CC}(i,j)$ (similar to $R_{MLO}(i,j)$).
Hence, the reinvented features are more friendly to compute the dual-view correlation loss, since each pixel is already encoded with its neighbors, among which the corresponding pixel in another view can find its align one.

For pixels in each row across views, the misalignment range $\sigma$ is hard to estimate, as implied in Fig.~\ref{fig:2}.
To tackle this, we also introduce a non-local attention mechanism to have each pixel contain the entire information of its belonging row, as shown in the right image of Fig.~\ref{fig:3}.
Combining both local and non-local attentions, the feature map extracted by the modified truncated ResNet101 is reinvented in a hybrid attention fashion.
In the next subsection, we instantiate the hybrid attention as our proposed hybrid attention module and give details of its two key constitutions, i.e., local relation block and non-local attention block.

\subsection{Hybrid Attention Module}
\label{sec:ii-c}

\begin{figure*}[tbp]	
	\centering	
	\includegraphics[width=\linewidth]{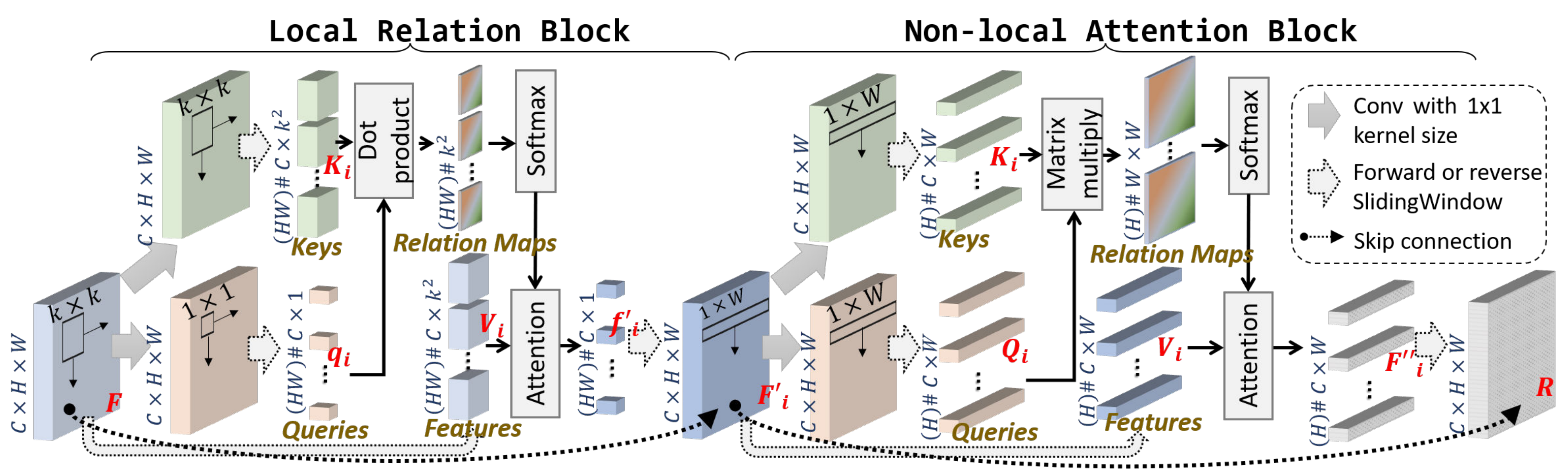}	
	\caption{Detailed architectures of two key blocks of our proposed DCHA-Net, i.e., local relation block and non-local attention block. $(HW)\#~C \times k^2$ means numerical units with the total number of $HW$ and the size of $C \times k^2$, and similar to others.}
	\label{fig:4}	
\end{figure*}

A vanilla attention block~\cite{vaswani2017attention} typically contains three numerical units, i.e., a query of feature $q \in \mathbb{R}^{C \times 1}$ to encode, features $V \in \mathbb{R}^{C \times D}$ to relate, and keys of features $K \in \mathbb{R}^{C \times D}$ to compute attentions with the query.
The relations are obtained by performing the dot products of the query with all keys, dividing each by $\sqrt{C}$, and applying a softmax function successively.
Hence, we have a new feature $f' \in \mathbb{R}^{C \times 1}$ encoded with all information from $V$ based on the attention block:
\begin{equation}
	\label{eq:3}
	f'^{T}=softmax(\frac{q^{T}K}{\sqrt{C}})V^{T}
\end{equation}

Next, we describe the two instantiations of Eq.~(\ref{eq:3}), i.e., local relation block and non-local attention block, which are designed to accomplish the manipulations visualized in Fig.~\ref{fig:3}.

\subsubsection{Local Relation Block}
As shown in the left part of Fig.~\ref{fig:4}, we first compute a key pool and a query pool from the original feature map $F \in \mathbb{R}^{C \times H \times W}$ (omitting the subscripts $CC$ and $MLO$) by two $1 \times 1$ convolution layers.
$C$ indicates the dimensions of the channel and $H \times W$ indicates the feature map size.

From the query pool denoted as the orange cuboid in Fig.~\ref{fig:4}, we extract feature vectors as queries at every location by a $1 \times 1$ sliding window (also can directly reshape), forming $HW$ queries $\{ q_i \in \mathbb{R}^{C \times 1}, i=0,...,HW-1 \}$.
Similarly, we can have keys $\{K_i \in \mathbb{R}^{C \times k^2},i=0,...,HW-1\}$ by performing a $k \times k$ sliding window with zero padding from the key pool denoted as the green cuboid in Fig.~\ref{fig:4} and by flattening.
The parameter $k$ controls the range of the possible misalignment, and we empirically set it to $3$ corresponding to a $24 \times 24$ receptive field in the original image $I$.

Each relation map is obtained by $softmax(q_{i}^{T}K_{i} / \sqrt{C})$, and thus those related features $\{f'_i \in \mathbb{R}^{C \times 1},i=0,...,HW-1\}$ can be calculated as:
\begin{equation}
	\label{eq:4}
	f_{i}'^{T}=softmax(\frac{q_i^{T}K_i}{\sqrt{C}})V_i^{T}
\end{equation}
where features $\{V_i \in \mathbb{R}^{C \times k^2},i=0,...,HW-1\}$ are extracted also by the sliding window and flattening on the original feature map $F$.
Those related features with the total number of $HW$ are at last packed together to $F' \in \mathbb{R}^{C \times H \times W}$ by the reverse sliding window process.
A skip connection is employed to have $F'=F' + F$.

By doing so, each entry of $F'$ carries information from its corresponding pixel in $F$, and neighboring ones defined by $V$.
Furthermore, the relation map defined based on $q_i$ and $K_i$ is also learned to make each entry in $F'$ pay more attention to its surroundings more likely to compensate misalignment to another view, and vice versa.

\subsubsection{Non-local Attention Block}
As shown in the right part of Fig.~\ref{fig:4}, we also compute a key pool and a query pool from $F'$ by another two $1 \times 1$ convolution layers.

Unlike the local relation block, we generate keys $\{K_i \in \mathbb{R}^{C \times W}\}$, queries $\{Q_i \in \mathbb{R}^{C \times W}\}$, and features $\{V_i \in \mathbb{R}^{C \times W}\}$ by a $1 \times W$ sliding window, where $i=0,...,H-1$, since we intend to relate each pixel to its belonging strip-like region (i.e., each row).
Note that, $Q_i$ is also equivalent to packing together all queries $q_i$ belonging to the same row.
Hence, the related features $\{F''_i \in \mathbb{R}^{C \times W},i=0,...,H-1\}$ can be calculated as:
\begin{equation}
	F_{i}''^{T}=softmax(\frac{Q_i^{T}K_i}{\sqrt{C}})V_i^{T}
\end{equation}

The overall attention-derived feature map $R$ is obtained by a reverse sliding window on $\{F''_i\}$.
Similarly, a skip connection is employed to have $R=R + F'$.

Empowered by the hybrid attention module, $R$ has each entry related to both its strip-like region and neighbors, and thus makes the dual-view correlation loss in Eq.~(\ref{eq:1}) play its full effect, better mining and preserving those underlying feature correlations across views without a need of registration.

\subsection{Loss for Training}
We contribute the proposed DCHA-Net to solve the holistic mammogram classification task.
To this end, we add two classification heads without weight sharing on the top of the extracted and reinvented feature maps, i.e., $R_{CC}$ and $R_{MLO}$ respectively, whose size is $1024 \times 32 \times 32$ as shown in Fig.~\ref{fig:1}.
Each head consisting of two layers first performs global average pooling to get a $1024$-d representation, and utilizes a full-connected layer and sigmoid function to predict a single unit $p$ indicating the possibility of the input image containing the malignant breast tumor lesion or not.
The classification losses for the two views are calculated as follows:
\begin{flalign}
	L_{clss}^{CC}&=-\left( ylog(p_{CC}) + (1-y)log(1-p_{CC}) \right) \\
	L_{clss}^{MLO}&=-\left(ylog(p_{MLO}) + (1-y)log(1-p_{MLO})\right)
\end{flalign}
where $p_{CC}$ and $p_{MLO}$ are predictions for CC and MLO view images respectively.
Totally, the final loss to train DCHA-Net is thus calculated as:
\begin{equation}
	L = L_{corr} + L_{clss}^{CC} +  L_{clss}^{MLO}
\end{equation}

\section{Experiments}
\subsection{Datasets and Experimental Setup}

\paragraph{INbreast:} The INbreast dataset collects in total 410 full-field digital mammographic images, from which 90 cases, i.e., patients, with both breasts and 25 cases with only one side of breast are included. Multiple different types of annotations are provided, including the BIRADS classification scores, mass/calcification masks for segmentation, and other annotations such as pectoralis muscles and distortions.

\paragraph{CBIS-DDSM:} The CBIS-DDSM dataset has in total 3071 scanned film mammography images (including 891 mass cases and 753 calcification cases). The CBIS-DDSM is a selected version of DDSM, with higher image and label quality, and more friendly access. It also contains precise annotations including ROI segmentation masks, bounding boxes and the BIRADS scores.

We label images as two classes: the BIRADS scores belonging to $\left\{1,2,3\right\}$ as normal or benign, and $\left\{4,5,6\right\}$ as malignant.
For INbreast, we randomly split 80$\%$ cases for training and the remaining 20$\%$ cases for test. For CBIS-DDSM, we follow its default division setting with 85$\%$ training cases and 15$\%$ test cases. Note that, no images from the same patient are cross-used in training and test sets for the two datasets.
During the training, we exclude cases with only single view, and augment the original data with random rotation and flipping.
For the evaluation, we utilize two metrics, i.e., the Accuracy and the Area Under the receiver operating characteristic Curve (ROC), i.e., the AUC value.
\begin{table*}[thbp]\footnotesize	
	\centering
	\begin{tabularx}{0.85\linewidth}{p{5.5cm}p{1.5cm}p{1.8cm}*2{X<{\centering\arraybackslash}}}
		\toprule[0.5px]
		Method & Views & Data Division & Accuracy & AUC \\
  		\midrule		
		Dataset: INbreast\cr
        \midrule[0.1px]
		Domingues \emph{et al.}~\cite{domingues2012inbreast} & Single & Image & 0.890 & - \\ 
        
		Pretrained CNN+RF~\cite{dhungel2016automated} & Single & Image & 0.910$\pm$0.02 & 0.760$\pm$0.23 \\ 
        
		Deep MIL~\cite{zhu2017deep} & Single & Image & 0.900$\pm$0.02 & 0.890$\pm$0.04 \\ 
        
		Shams \emph{et al.}~\cite{shams2018deep} & Single & Image  & 0.935$\pm$0.03 & 0.925$\pm$0.02 \\ 
        
		RGP~\cite{shu2020deep} & Single & Image & 0.919$\pm$0.03 & 0.934$\pm$0.03 \\ 
        
		GGP~\cite{shu2020deep} & Single & Image & 0.922$\pm$0.02 & 0.924$\pm$0.03 \\
        
		Carneiro \emph{et al.}~\cite{carneiro2017automated} & Dual & Patient & - & 0.860$\pm$0.09 \\
  
		MCRLA ~\cite{li2021deep} & Dual & Patient & 0.912 & 0.942 \\
  
		\textbf{DCHA-Net}& Dual & Patient & \textbf{0.955$\pm$0.01} & \textbf{0.950$\pm$0.02} \\ 
  		\midrule		
		Dataset: CBIS-DDSM\cr  
        \midrule[0.1px]
  		Deep MIL~\cite{zhu2017deep} & Single & Patient & 0.742$\pm$0.03 & 0.791$\pm$0.02 \\ 
        
		RGP~\cite{shu2020deep} & Single & Patient & 0.762$\pm$0.02 & 0.838$\pm$0.01 \\ 
        
		GGP~\cite{shu2020deep} & Single & Patient & 0.767$\pm$0.02 & 0.823$\pm$0.02 \\
  
		MCRLA~\cite{li2021deep} & Dual & Patient & 0.766 & 0.824 \\
  
		Petrini \emph{et al.}*~\cite{petrini2022breast} & Dual & Patient & - & 0.842$\pm$0.03 \\  
        
		\textbf{DCHA-Net}& Dual & Patient & \textbf{0.781$\pm$0.01} & \textbf{0.846$\pm$0.01} \\ 
    \bottomrule[0.5px]
	\end{tabularx}
 \caption{Quantitative comparison of different methods on both the INbreast and the CBIS-DDSM dataset. Our final results of DCHA-Net are obtained by using data-augmentation techniques during training. Results of other groups except ours are directly inherited from papers of Shams \emph{et al.}, RGP/GGP, Carneiro \emph{et al.}, MCRLA and Petrini \emph{et al.}. `*' indicates results without test-time augmentation for a fair comparison.}\label{table1}
\end{table*}

\subsection{Implementation Details}
\label{sec:iii-b}
For data pre-processing, we first use OpenCV edge detection to remove background. We use the provided GT masks to remove pectoralis muscles for INbreast. We manually find a line fitting the chest wall, and then remove regions on the non-breast side for CBIS-DDSM. We use lines to fit chest wall and align the two lines across views.

The proposed DCHA-Net is implemented with the Pytorch library and trained on one NVIDIA GeForce RTX 3090 GPU with 24 GB memory. 
We use a pretrained ResNet-101 for weight initialization, and an Adam optimizer.
The learning rate starts at $5e$-$5$ and gradually decays by 0.9.
Our method uses two classification heads and thus gives two predictions for a dual-view image of a breast, and we take the average output unit as the final predicted probability of being malignant and binarize it using a threshold 0.5.

\subsection{Comparison with the State-of-the-arts}
We first compare our method with eight previous state-of-the-arts on INbreast, including Domingues \emph{et al.}~\cite{domingues2012inbreast}, deep MIL~\cite{zhu2017deep}, Shams \emph{et al.}~\cite{shams2018deep}, RGP and GGP~\cite{shu2020deep}, Carneiro \emph{et al.}~\cite{carneiro2017automated} and MCRLA~\cite{li2021deep}.
As shown in Table~\ref{table1}, most approaches are single view-based methods, which can hardly achieve satisfactory performance without extracting and utilizing dual-view information. 
Moreover, deep MIL, Shams \emph{et al.} and RGP/GGP all divide data by images, where images in the same case may be split into the training and testing set at the same time. 
In contrast, DCHA-Net effectively mines dual-view features on patient-division data, which surpasses all state-of-the-art approaches by a great margin, achieving the best results in terms of the average Accuracy ($0.955$) and the AUC ($0.950$).

We also evaluate on the CBIS-DDSM dataset. 
As shown in the bottom of Table~\ref{table1}, we compare our DCHA-Net with Deep MIL, RGP/GGP, MCRLA and Petrini \emph{et al.}~\cite{petrini2022breast}. 
Note that all approaches followed the default data division in the CBIS-DDSM dataset and trained on patient-division data. 
In comparison, our approach remarkably achieves the best average accuracy and the AUC value, remaining consistency results as those on INbreast.

Comparing results between the two datasets, it is worth noting that the performance on the INbreast dataset is greater than that on the CBIS-DDSM dataset.
We believe this is possibly caused by different image quality. 
For instance, the INbreast images were collected using more advanced mammography screening techniques, which can help extract more useful features during training.
        
        
        
\subsection{Ablation Analysis}

\subsubsection{Ablation Analysis of Key Components}

We conduct an ablation study on the INbreast dataset to analysis impacts of key components in the DCHA-Net.
Here, we disable the data augmentation techniques used in comparison with the state-of-the-arts.
Table~\ref{table2} shows the comparison results of six variants, including: 
1) “Baseline” which directly trains on clear ResNet backbones (see the $1^{st}$ row); 
2) “Corr. only” that only utilizes correlation constraints during training (the $2^{nd}$ row); 
3) “Corr. plus local relation” that uses only local attention with correlation constraints (the $3^{rd}$ row); 
4) “Corr. plus non-local atten.” that uses only non-local attention with correlation constraints (the $4^{th}$ row);
5) “Hybrid-atten. only” that only adds the hybrid-attention module (the $5^{th}$ row) and 
6) “DCHA-Net” that utilizes both the hybrid-attention module and correlation constraints (the $6^{th}$ row).

Four observations can be made from the results. First, the baseline achieves the worst performance, indicating the significance of both the correlation constraints and the hybrid-attention module. More specifically, the correlation constraints and the hybrid-attention module can respectively result in an increment of $1.136\%$ and $1.515\%$ in the average accuracy, and an increment of $0.007$ and $0.043$ in the AUC value. Second, compared to Baseline, DCHA-Net greatly improves the accuracy by $4.924\%$ and the AUC by $0.067$.
Third, when only using local
(non-local) attention with correlation maximization, ACC is
$90.152\%$ ($89.773\%$) and AUC is 0.909 (0.901), inferior to
those achieved by combining both attentions (i.e., hybrid). Fourth, we can observe that the correlation constraints can result in higher improvements after using the hybrid-attention module.
This indicates that the hybrid-attention module can effectively tackle the dual-view spatial misalignment problem, and help capture correct correlation information maximally. 

In addition, we also perform student t-test, and report p-values for both metrics.
Comparing DCHA-Net with Baseline (Correlation-only), the p-values for Accuracy and AUC are $7.78e$-$4$ and $2.81e$-$2$ ($1.11e$-$2$ and $1.05e$-$3$), respectively.
Comparing Corr. only with Baseline, the corresponding p-values are $2.51e$-$1$ and $7.66e$-$1$.
This indicates the effectiveness of the hybrid-attention module on misalignment.
\begin{figure}[th]	
	\centering	
	\includegraphics[width=0.89\linewidth]{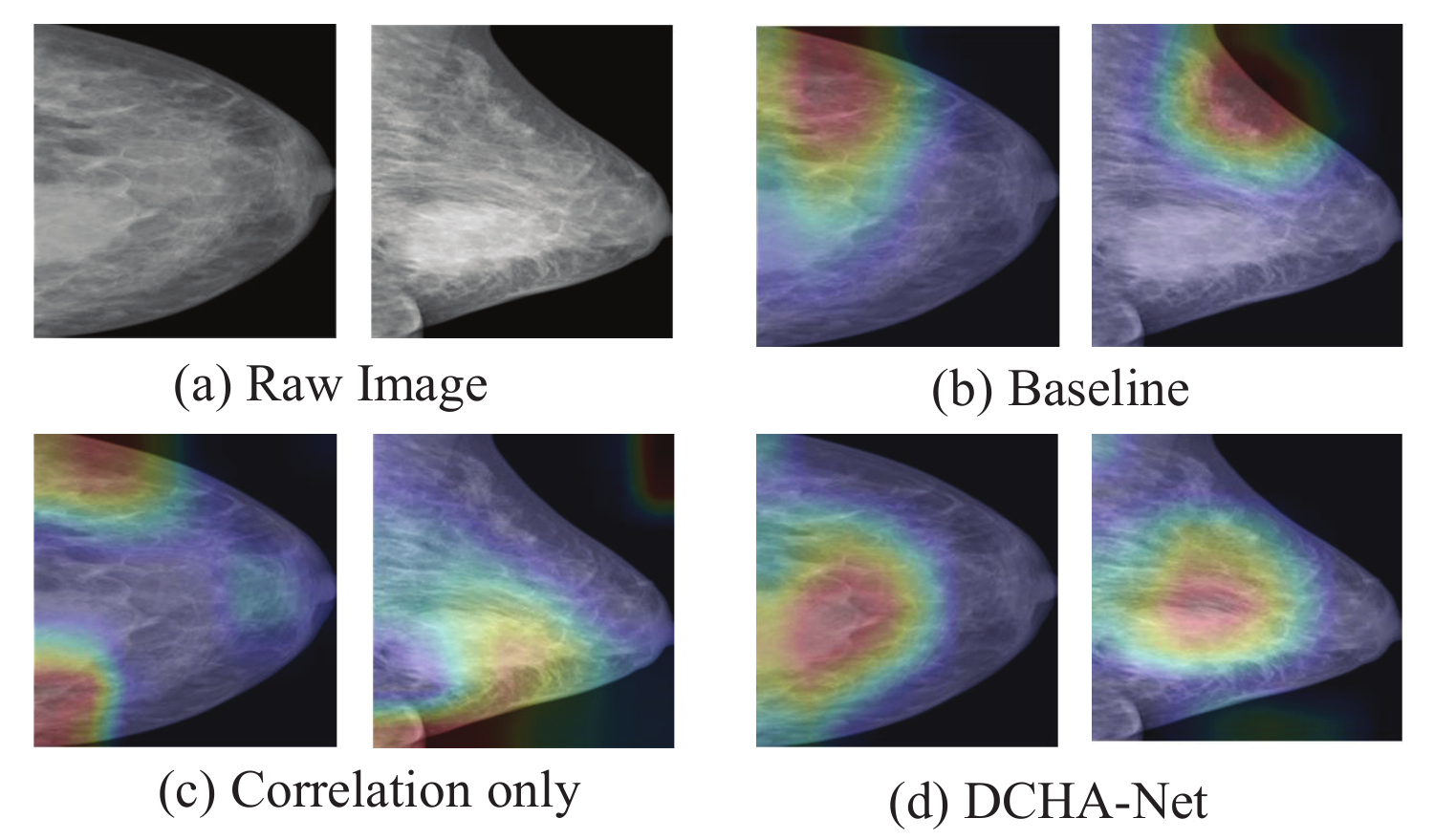}
	\caption{Visualization of the most suspicious malignant areas predicted by different methods. Each group is sampled from a same case with paired CC and MLO images.}
	\label{fig:5}	
\end{figure}
\begin{table}\small
	\centering
	\begin{tabularx}{\linewidth}{p{1.5cm}p{1.5cm}p{1.5cm}p{1.2cm}<{\centering\arraybackslash}p{0.8cm}<{\centering\arraybackslash}}
		\toprule[0.5px]
		Local atten. & Non-local atten. & Dual-view corr. & Acc. (\%) & AUC \\ \midrule[0.1px]
        \textcolor[rgb]{0.8,0,0.2}{\XSolidBrush} & \textcolor[rgb]{0.8,0,0.2}{\XSolidBrush} & \textcolor[rgb]{0.8,0,0.2}{\XSolidBrush} & 87.879 & 0.870\\ 
        \textcolor[rgb]{0.8,0,0.2}{\XSolidBrush} & \textcolor[rgb]{0.8,0,0.2}{\XSolidBrush} & \textcolor[rgb]{0,0.8,0.2}{\Checkmark} & 89.015 & 0.877\\ 
        \textcolor[rgb]{0,0.8,0.2}{\Checkmark} & \textcolor[rgb]{0.8,0,0.2}{\XSolidBrush} & \textcolor[rgb]{0,0.8,0.2}{\Checkmark} & 90.152 & 0.909\\ 
        \textcolor[rgb]{0.8,0,0.2}{\XSolidBrush} & \textcolor[rgb]{0,0.8,0.2}{\Checkmark} & \textcolor[rgb]{0,0.8,0.2}{\Checkmark} & 89.773 & 0.901\\
        \textcolor[rgb]{0,0.8,0.2}{\Checkmark} & \textcolor[rgb]{0,0.8,0.2}{\Checkmark} & \textcolor[rgb]{0.8,0,0.2}{\XSolidBrush} & 89.394 & 0.913\\ 
        \textcolor[rgb]{0,0.8,0.2}{\Checkmark} & \textcolor[rgb]{0,0.8,0.2}{\Checkmark} & \textcolor[rgb]{0,0.8,0.2}{\Checkmark} & \textbf{92.803}  &\textbf{0.937}\\ 
        \bottomrule[0.5px]
	\end{tabularx}
\caption{Ablation analysis of the proposed hybrid-attetion module and employed correlation maximization.}\label{table2}
\end{table}

To further demonstrate the improvements of mining dual-view information, we used Grad-CAM~\cite{selvaraju2017grad} to visualize the most suspicious malignant areas (e.g., mass) predicted by different groups of methods. Grad-CAM uses the gradient as weight to highlight attentive areas, which are those more contribute to the final classification prediction. Visual sign of lesions is mostly related to classification label.
As shown in Fig.~\ref{fig:5}(b), without correlation loss, the gradient cannot flow across views, making the two view features unable to “cross-check” and easily distracted by some lesion-irrelevant regions. 
Therefore, the baseline model leans to focus on lesion-irrelative and cross-view mismatched areas. 
With correlation only, the two view features cannot “cross-check” at the truly matched places, bringing incorrect and confusing information from another view due to the spatial misalignment problem (see Fig.~\ref{fig:5}(c)).
As shown in Fig.~\ref{fig:5}(d), with correlation plus hybrid-attention, the misalignment is alleviated in the feature space, and the matched lesion-relevant regions can be successfully highlighted by Grad-CAM. 

\subsubsection{Effectiveness of Hybrid-attention Module}
The hybrid-attention module consists of two basic blocks, i.e., the local attention block and the non-local attention block. 
We conduct an ablation study of their effectiveness on INbreast using two different settings, i.e., ``mixed views'' and ``single view'', and the data augmentation techniques are also disabled.
Under ``mixed views'', we mix images from the two views together and simply train our model by using a single shared classification head without constraining the dual-view correlation.
Under ``single view'', we split the data into two parts, and each contains images from a single view.
We independently train a model for each part, and thus the correlation constraint is naturally disabled.
\begin{table}\small
	\centering
	\begin{tabularx}{\linewidth}{p{1.3cm}*2{>{\centering\arraybackslash}X}*2{p{1.1cm}<{\centering\arraybackslash}}}
		\toprule[0.5px]
		Training view & Local Relation Block& Non-Local Attention Block & Accuracy ($\%$) on CC view & Accuracy ($\%$) on MLO view \\ \midrule[0.1px]
		\multirow{4}{1cm}{Mixed views} & \textcolor[rgb]{0.8,0,0.2}{\XSolidBrush} & \textcolor[rgb]{0.8,0,0.2}{\XSolidBrush} & 88.636 & 87.879 \\ 
		& \textcolor[rgb]{0,0.8,0.2}{\Checkmark} & \textcolor[rgb]{0.8,0,0.2}{\XSolidBrush} & 89.773 & 88.636 \\ 
		& \textcolor[rgb]{0.8,0,0.2}{\XSolidBrush} & \textcolor[rgb]{0,0.8,0.2}{\Checkmark} & 89.394 & 89.394 \\ 
		& \textcolor[rgb]{0,0.8,0.2}{\Checkmark} & \textcolor[rgb]{0,0.8,0.2}{\Checkmark} & \textbf{90.152} & \textbf{90.909} \\ \midrule[0.1px]
		\multirow{4}{1cm}{Single view} & \textcolor[rgb]{0.8,0,0.2}{\XSolidBrush} & \textcolor[rgb]{0.8,0,0.2}{\XSolidBrush} & 88.636 & 87.121 \\ 
		& \textcolor[rgb]{0,0.8,0.2}{\Checkmark} & \textcolor[rgb]{0.8,0,0.2}{\XSolidBrush} & 89.394 & 87.879 \\ 
		& \textcolor[rgb]{0.8,0,0.2}{\XSolidBrush} & \textcolor[rgb]{0,0.8,0.2}{\Checkmark} & 89.394 & 87.879 \\ 
		& \textcolor[rgb]{0,0.8,0.2}{\Checkmark} & \textcolor[rgb]{0,0.8,0.2}{\Checkmark} & \textbf{90.152} & \textbf{88.636} \\ \bottomrule[0.5px]
	\end{tabularx}
\caption{Ablation study of the effectiveness of the two different attention blocks in hybrid attention module under two different settings of images for training. The correlation maximization is disabled for verifying the hybrid attention purely.}\label{table3}
\end{table}

We report the performance on each view for the two settings, and the results are shown in Table~\ref{table3}. As can be seen from the results, both the local-attention block and the non-local attention block can contribute to large improvements.
For instance, under ``mixed views'', the local-attention block and the non-local attention block respectively result in an increment of $0.947\%$ and $1.137\%$ in the average accuracy. In conjunction of these two components, the hybrid-attention module achieves the best performance, e.g., improving the average accuracy by $2.273\%$ and $1.516\%$ respectively under these two settings. This also implies that the two components can work synthetically to result in a more robust performance. 
\section{Conclusion}
In this paper, we propose a novel end-to-end DCHA-Net which contains two key components for robust holistic mammographic classification.
First, the dual-view correlation loss aims at maximizing paired feature similarity across two views, which effectively helps capture consistent and complementary information for better mammographic classification accuracy.
In addition, the hybrid-attention module reinvents information from local and strip-like non-local regions into every pixel, alleviating negative influences brought by the spatial misalignment problem and guaranteeing the extracted dual-view correlated features correct.
Extensive experimental results on both the INbreast and CBIS-DDSM datasets demonstrate that our proposed DCHA-Net can significantly improve the breast cancer diagnosis performance and outperform previous state-of-the-art methods.

\section*{Contribution Statement}
Zhiwei Wang and Junlin Xian are the
co-first authors contributing equally to this work. Qiang Li and Xin Yang are corresponding authors.

\section*{Acknowledgements}
This work was supported in part by National Natural Science Foundation of China (Grant No. 62202189), Fundamental
Research Funds for the Central Universities (2021XXJS033), research grants from Wuhan United Imaging Healthcare Surgical Technology Co., Ltd.

\bibliographystyle{named}
\bibliography{ijcai23}

\end{document}